\documentclass[11pt]{article}

\usepackage{graphicx}
\usepackage{amsmath}
\usepackage{url}
\usepackage{hyperref}
\usepackage{booktabs}
\usepackage[margin=1in]{geometry}
\usepackage{multirow}
\usepackage{enumitem}
\usepackage{xcolor}
\usepackage{svg}
\usepackage{tikz}
\usepackage{subcaption}

\title{\textbf{SynGP500: A Clinically-Grounded Synthetic Dataset of Australian General Practice Medical Notes}}

\author{
    Piyawoot Songsiritat, MBBS, FRACGP, B.Eng (Computer Engineering)\\
    General Practitioner \& Independent Clinical NLP Researcher\\
    \texttt{piyawoot.song [at] gmail.com}
}

\date{2025}

\begin{document}
\maketitle

\noindent\textit{Preprint}

\begin{abstract}
We introduce \textbf{SynGP500}, a clinician-curated collection of 500 synthetic Australian general practice medical notes. The dataset integrates curriculum-based clinical breadth (RACGP 2022 Curriculum), epidemiologically-calibrated prevalence (BEACH study), and diverse consultation contexts. This approach systematically includes both common presentations and less-common curriculum-specified conditions that GPs must recognize but appear infrequently in single-practice populations, potentially supporting more generalizable model training than datasets constrained by naturally occurring case distributions.

SynGP500 is messy by design, reflecting the authentic complexity of healthcare delivery: telegraphic documentation, typos, patient non-adherence, socioeconomic barriers, and clinician-patient disagreements—unlike sanitized synthetic datasets that obscure clinical realities.

Multi-faceted validation demonstrates dataset quality through epidemiological alignment with real Australian GP consultation patterns (BEACH study), stylometric analysis confirming high linguistic variation, semantic diversity analysis demonstrating broad coverage, and exploratory downstream evaluation using self-supervised medical concept extraction, showing F1 improvements.

SynGP500 addresses a critical national gap, providing researchers and educators with a resource for developing and evaluating clinical NLP methods for Australian general practice while inherently protecting patient privacy.
\end{abstract}

\section{Introduction}

Publicly available clinical text data for Australian general practice -- where most healthcare is delivered -- is essentially non-existent. Dispersed practice ownership, absence of a national research repository, and strict privacy requirements severely limit data availability, creating substantial barriers to developing NLP systems that can support clinical decision-making, quality improvement, and research in this critical healthcare setting.

The importance of Australian-specific datasets for primary care NLP cannot be overstated. Australian general practice has distinctive characteristics that existing datasets do not capture: terminology differences (e.g., Australian ``Panadol'' vs U.S. ``Tylenol''), medication availability through the Pharmaceutical Benefits Scheme (PBS), clinical guidelines from RACGP and Therapeutic Guidelines, consultation patterns documented in the BEACH epidemiological study, and the unique challenges of delivering care across Australia's diverse geographic and socioeconomic contexts.

We introduce \textbf{SynGP500}, a clinician-curated collection of 500 synthetic Australian general practice medical notes designed to capture authentic clinical complexity for NLP research and education. Our contribution includes:

\begin{itemize}[noitemsep]
    \item \textbf{To our knowledge, the first publicly available Australian general practice clinical text corpus} that is synthetic, privacy-preserving, and accessible without ethics approval—addressing a critical national infrastructure gap for clinical NLP research and education
    \item \textbf{Curriculum-based framework aiming to address the long-tail problem:} RACGP 2022 Curriculum for clinical breadth, BEACH epidemiological calibration for realistic prevalence, and SNOMED CT-AU ontology mapping—ensuring systematic coverage including less-common presentations that appear infrequently in single-practice populations. Framework scalable to rare conditions via specialty curriculum (Table~\ref{tab:coverage-tiers})
    \item \textbf{Authentic practice complexity:} nine consultation contexts (standard clinic, telehealth, residential aged care, bulk-billing, home visits, after-hours, Aboriginal health services, community health, mobile outreach) across metropolitan to remote settings (MM1--MM7), with varied clinician documentation styles and psychosocial complexity (cultural factors, adherence challenges, unrealistic expectations) capturing realistic clinical and linguistic diversity
    \item \textbf{Multi-faceted validation} through epidemiological calibration (BEACH study comparison showing close alignment with real consultation patterns), stylometric analysis demonstrating high linguistic variation, semantic diversity analysis demonstrating absence of mode collapse, and downstream task evaluation indicating practical utility for clinical concept extraction
    \item \textbf{Public release} under CC BY-NC-SA 4.0 license supporting research, prototyping, and education without ethics approval barriers
\end{itemize}

\section{Related Work}

Existing clinical text datasets do not address primary care contexts, particularly Australian general practice. Large-scale corpora such as MIMIC-III/IV \cite{mimic} and i2b2 \cite{i2b2} focus on U.S. hospital inpatient and ICU settings rather than primary care. While some nations maintain research databases of primary care records (e.g., UK's Clinical Practice Research Datalink, Netherlands IPCI database), these remain access-restricted, geographically-specific, and do not address Australian clinical contexts. Primary care documentation presents distinctive NLP challenges compared to hospital settings: broader clinical scope spanning minor acute illness to complex multimorbidity, time-constrained telegraphic documentation, and the long-tail problem where less-common presentations collectively form a substantial part of practice.

While synthetic data generation offers a privacy-preserving alternative for healthcare NLP \cite{tucker2020, chen2021, synthetic}, existing approaches have limitations. Synthea \cite{synthea} generates structured EHR data but not free-text clinical notes. Recent advances in large language models enable synthetic clinical text generation, but naïve LLM generation suffers from distributional collapse, producing templated, low-diversity notes when generating many cases. Our methodology counters this via structured conditioning across clinical breadth, prevalence, and contextual variation (Sections~3.1--3.2), validated through semantic diversity analysis (Section~5.2). To our knowledge, no publicly available corpus of primary care clinical text for Australian general practice—real or synthetic—currently exists.

\section{Methods}

The dataset was constructed through curriculum-based condition selection, multi-dimensional grounding, synthetic note generation, and LLM-assisted clinician validation.

\subsection{Case Selection}

Case selection balanced epidemiological realism with clinical breadth. Table~\ref{tab:coverage-tiers} shows the three coverage tiers.

\begin{table}[ht]
\centering
\caption{Prevalence and curriculum-based stratification of medical conditions}
\label{tab:coverage-tiers}
\begin{tabular}{p{3.5cm}p{4cm}p{6cm}}
\toprule
\textbf{Coverage Tier} & \textbf{Selection Approach} & \textbf{Example Conditions} \\
\midrule
\textbf{Common} & RACGP curriculum & Hypertension, Type 2 diabetes, Upper respiratory tract infection \\
\addlinespace
\textbf{Less common but important} & RACGP curriculum & Addison's disease, Diabetic ketoacidosis, Giant cell arteritis \\
\addlinespace
\textbf{Rare (long-tail)} & Specialty curriculum and literature & Granulomatosis with polyangiitis, Dyskeratosis congenita, Takayasu's arteritis \\
\bottomrule
\end{tabular}
\vspace{0.3cm}

\noindent\small\textit{SynGP500's 500-note release addresses Common + Less common tiers through RACGP curriculum. The framework can scale to rare long-tail conditions via specialty curriculum and literature.}
\end{table}

\subsubsection{BEACH Epidemiological Calibration}

To ensure realistic condition prevalence, case selection was weighted using BEACH study \cite{beach} data—a continuous national study of Australian general practice activity—matching the dataset's frequency distribution to real-world presentations seen by Australian GPs.

\subsubsection{RACGP Curriculum Alignment}

The RACGP 2022 Curriculum \cite{racgp} ensured systematic coverage across acute, chronic, mental health, preventive, dermatological, and multimorbidity domains. This curriculum-driven approach ensures coverage of less common but clinically important presentations that appear too infrequently in individual practices to support robust training, partially addressing the long-tail problem (truly rare conditions remain future work). This breadth is hypothesized to reduce domain shift when deployed in real general practice settings where GPs must handle both common and uncommon presentations.

\subsection{Multi-Dimensional Grounding}

Cases were grounded across five dimensions:

\subsubsection{Clinical Guidelines}
Clinical reasoning was informed by the author's curated collection of contemporary Australian clinical guidelines and prescribing standards, which provided the evidence-based foundation for synthetic clinical decision-making.

\subsubsection{Consultation Context Variation}
Nine consultation settings were incorporated with setting-appropriate clinical constraints: standard clinic, bulk-billing (high volume, time pressure), residential aged care (multimorbidity), telehealth (examination limitations), home visits, after-hours (limited resources), community health, Aboriginal health services (cultural safety), and mobile outreach (resource constraints). Setting-appropriate variation in documentation style, investigation accessibility, and management pragmatism reflects real-world general practice diversity.

\subsubsection{Remoteness Adaptation}
Management plans were adapted to remoteness classifications (MM1--MM7, Modified Monash Model from metropolitan to very remote), increasing scenario diversity and clinical realism. For example, STEMI (major heart attack) management in MM1 areas reflects immediate access to PCI (stenting), while MM6--MM7 remote settings require thrombolysis (clot buster medication), patient retrieval, and fundamentally different treatment decisions shaped by resource availability.

\subsubsection{Psychosocial Complexity}
Psychosocial factors reflecting social determinants of health were systematically integrated: housing instability (affecting medication storage, follow-up reliability); cultural factors (influencing health beliefs, treatment preferences); language barriers (requiring interpreters, affecting health literacy); family dynamics (impacting care support, treatment adherence); and adherence challenges (financial constraints, work schedules, health literacy). These factors actively influence documented clinical reasoning—for example, a patient's casual employment affecting their ability to attend physiotherapy, or housing instability prompting GP consideration of injectable rather than oral antibiotics—rather than appearing as isolated demographic details.

\subsection{Clinical Note Generation}

\subsubsection{Generation Method}
Notes were generated through the multi-dimensional grounding framework (Sections 3.1--3.2) using GPT-5 (OpenAI, batch API mode, temperature 1.0) as the text generation tool. Recent work demonstrates that large language models encode substantial clinical knowledge \cite{singhal2023}, motivating their use for clinically grounded text generation.

Generated notes span realistic length variation—from telegraphic to detailed—with note length reflecting case complexity, ensuring even brief notes contain meaningful clinical content and prioritizing substantiveness over matching the sparsest extremes of time-pressured documentation.

\subsubsection{Clinician Personas}
A library of synthetic GP writing personas was developed to capture authentic variation in clinical documentation patterns. Personas varied across multiple dimensions: verbosity (from telegraphic to expansive); abbreviation frequency and style; documentation structure (SOAP format vs narrative vs hybrid); clinical reasoning explicitness (detailed differential diagnosis vs direct impression); safety-netting detail; and stylistic idiosyncrasies including natural typo rates.

This variation is evident in the illustrative examples presented in Section~6: the impetigo case (Section~6.1) demonstrates extreme abbreviation and time-pressure documentation; the prepatellar bursitis case (Section~6.2) shows moderate detail with pragmatic problem-oriented structure; while the ankle sprain case (Section~6.3) exhibits more defensive documentation explicitly recording patient disagreement and clinical justification for withholding requested investigations.

These persona variations ensure that NLP systems trained on SynGP500 encounter the realistic documentation diversity found in actual general practice rather than homogeneous, template-driven text—preparing models for real-world deployment where clinicians exhibit substantial individual variation in writing style and clinical reasoning documentation.

\subsection{Validation Pipeline}

Generated notes underwent LLM-assisted screening to flag potential serious errors—biological impossibilities, medications with outright wrong indications, or factually incorrect clinical content—followed by manual review and correction of confirmed errors by the author (Australian GP). Legitimate variations in clinical practice and documentation style were preserved.

\section{Dataset Characteristics}

SynGP500 comprises 500 synthetic Australian general practice medical notes with the following characteristics:

\begin{itemize}[noitemsep]
    \item \textbf{Note length:} Average 606 $\pm$ 257 words (range 213--1,444 words)
    \item \textbf{Total corpus size:} 245,271 words
    \item \textbf{Patient population:} Adult and elderly patients (18+ years); pediatric cases not included in this release
    \item \textbf{Consultation settings:} Nine contexts—standard GP clinic, bulk billing clinic, residential aged care, telehealth, home visits, after-hours, community health, Aboriginal health services, mobile outreach
    \item \textbf{Geographic distribution:} Metropolitan, regional, rural, and remote locations (MM1--MM7 remoteness classifications)
    \item \textbf{Clinical conditions:} Aligned with RACGP registrar curriculum; epidemiologically calibrated to BEACH study prevalence patterns
    \item \textbf{SNOMED CT-AU coverage:} All cases mapped to SNOMED CT-AU concepts
    \item \textbf{Documentation styles:} Multiple synthetic clinician personas with varying verbosity, abbreviation patterns, note structure, and typo rates
\end{itemize}

The dataset is distributed as plain text files (UTF-8 encoding) with filenames following the pattern: \texttt{\{SNOMED\_code\}\_\{ID\}\_\{condition\_name\}.txt}

\section{Dataset Validation}

Direct comparison to real GP clinical notes would provide the strongest validation but requires ethics approval not obtained for this initial release. Our validation strategy therefore combines three complementary approaches: (1) comparison against publicly available statistics (BEACH study) to validate distributional realism; (2) stylometric and semantic diversity analyses to confirm absence of mode collapse and template artifacts common in synthetic datasets; and (3) downstream NER evaluation to demonstrate practical utility. Formal comparison with real GP documentation is planned pending ethics approval.

\subsection{Stylometric Analysis}

Comprehensive stylometric analysis confirms that SynGP500 exhibits high linguistic variation and diversity rather than template-driven text generation:

\begin{itemize}[noitemsep]
    \item \textbf{Lexical diversity (MATTR) \cite{covington2010}:} 0.946 (25-word window), 0.909 (50-word), 0.858 (100-word)—indicating high vocabulary richness and genuine linguistic variation rather than repetitive patterns
    \item \textbf{Natural imperfections:} 0.83\% typo rate (2,032 typos across 245,271 words)—reflecting realistic documentation errors during time-pressured consultations
    \item \textbf{Length variation:} Average 606 $\pm$ 257 words per note (range 213--1,444), CV 0.42--0.47—substantial diversity reflecting that different clinical encounters require different documentation detail
    \item \textbf{Style variability:} Article density CV 0.84, copula density CV 1.19—high inter-note variation indicating diverse clinician writing styles rather than rigid templates
    \item \textbf{Medical authenticity:} 48.3\% medical term density—appropriate clinical content concentration
\end{itemize}

These metrics validate that the dataset captures authentic stylistic variation rather than homogeneous, artificially polished text.

\subsection{Semantic Diversity Analysis}

Note level embeddings (all-mpnet-base-v2 \cite{song2020mpnet,reimers2019sentence}, averaged from first and last 512 tokens for notes exceeding model context) demonstrate broad semantic diversity. Pairwise document cosine similarity (mean 0.52, range 0.09--0.95) and UMAP dimensionality reduction \cite{mcinnes2018umap} across the embedding space provide evidence against mode collapse typical of naïve LLM generation.

\begin{figure}[ht]
    \centering
    \begin{subfigure}[b]{0.48\linewidth}
        \centering
        \includegraphics[width=\linewidth]{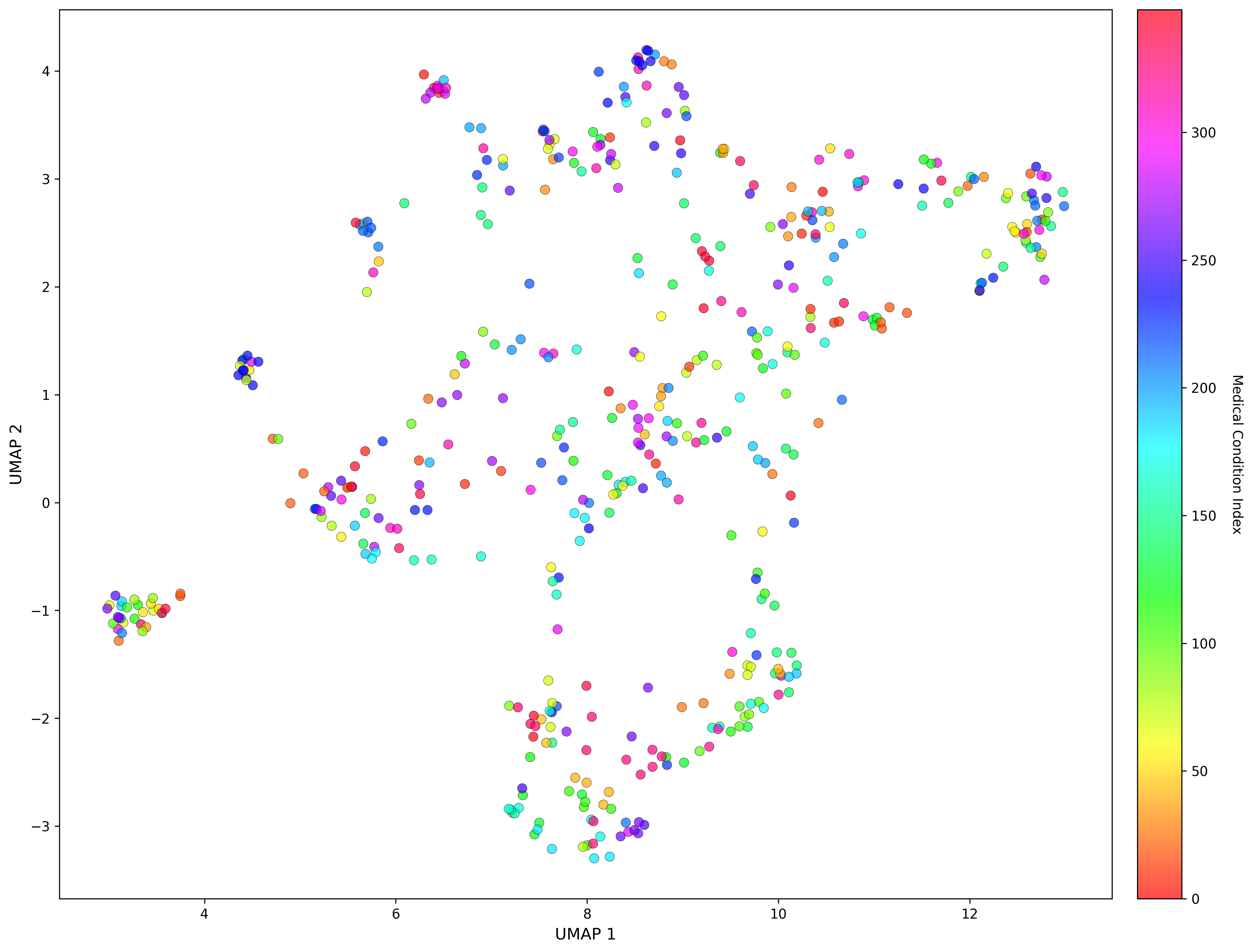}
        \caption{Clinical content}
        \label{fig:umap_clinical}
    \end{subfigure}
    \hfill
    \begin{subfigure}[b]{0.48\linewidth}
        \centering
        \includegraphics[width=\linewidth]{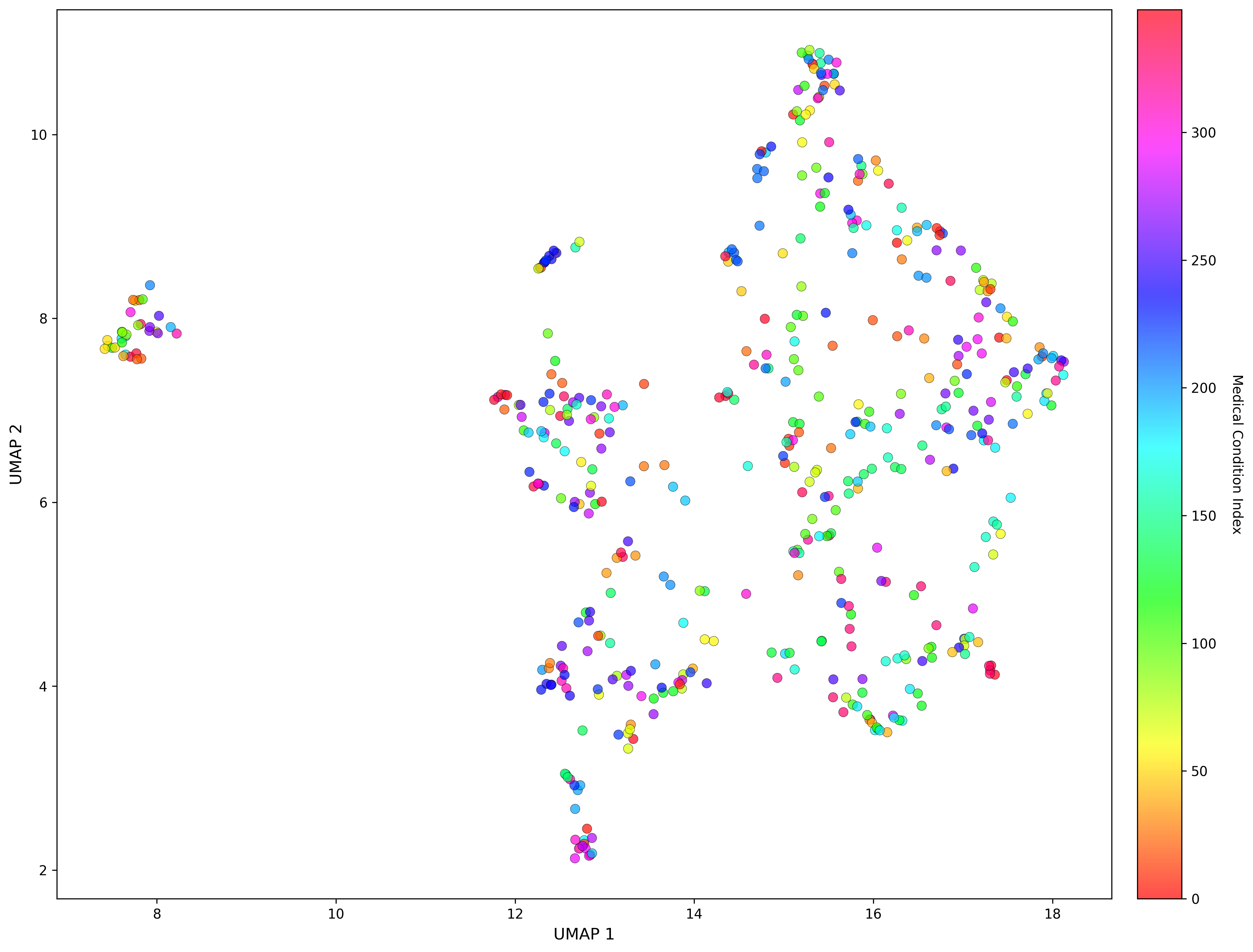}
        \caption{With persona writing style applied}
        \label{fig:umap_styled}
    \end{subfigure}
    \caption{UMAP projection of SynGP500 embeddings colored by medical conditions.}
    \label{fig:umap}
\end{figure}

\subsection{Epidemiological Validation}

Dataset distribution was assessed through LLM-based categorization into BEACH presenting complaint categories, with manual review of 10\% confirming classification accuracy. Table~\ref{tab:beach} shows representative examples demonstrating strong alignment with real-world Australian GP consultation patterns, with most categories matching within $\pm$1--2\% of BEACH prevalence (full 28-category comparison in Appendix~A). The lower ``Other'' category (-12.3\%) reflects intentional emphasis on RACGP curriculum-relevant conditions for enhanced educational and research utility while maintaining overall epidemiological realism.

\begin{table}[h]
\centering
\footnotesize
\setlength{\tabcolsep}{4pt}
\caption{Representative comparison of SynGP500 vs BEACH epidemiology (full table in Appendix~A)}
\begin{tabular}{lccc}
\toprule
\textbf{Condition Category} & \textbf{BEACH (\%)} & \textbf{SynGP500 (\%)} & \textbf{Difference} \\
\midrule
Prescription & 8.8 & 7.2 & –1.6 \\
Check-up & 8.1 & 6.4 & –1.7 \\
Test results & 6.7 & 7.0 & +0.3 \\
Cough & 4.1 & 3.6 & –0.5 \\
Depression & 1.3 & 2.6 & +1.3 \\
Headache & 1.1 & 2.0 & +0.9 \\
Anxiety & 1.0 & 1.2 & +0.2 \\
Sleep disturbance & 0.7 & 1.2 & +0.5 \\
Other & 40.4 & 28.1 & –12.3 \\
\bottomrule
\multicolumn{4}{l}{\footnotesize Most categories within $\pm$1--2\%. Complete 28-category comparison in Appendix~A.}
\end{tabular}
\label{tab:beach}
\end{table}

\subsection{Medical Concept Extraction Evaluation}

\subsubsection{Model}
We used MedCAT \cite{medcat} v2.0.0-dev (commit a7d2a337) with SNOMED CT-AU \cite{snomed} vocabulary from the Australian National Clinical Terminology Service (NCTS SCT RF2 Distribution 32506021000036107, August 2025 snapshot) and the pre-built MedMentions vocab file from the MedCAT repository. The Clinical Database (CDB) was constructed from 366,306 active SNOMED concepts filtered to 16 clinically-relevant semantic types (e.g., disorder, finding, procedure, substance), augmented with 664 author-curated common Australian medical acronyms mapped to SNOMED concepts. Configuration: remove\_parenthesis=5, min\_letters\_required=2. FSNs and synonyms map to identical concept IDs, enabling ontology-grounded concept matching.

\subsubsection{Evaluation Set}
Access to real Australian GP documentation requires ethics approval, which was not obtained for this initial release. To enable pragmatic validation, the author created 19 clinician-authored fictional GP notes representing a typical workday in Australian general practice, containing 648 manually annotated entities using a web-based tool supporting span selection and SNOMED CT-AU concept search. While this approach has limitations (fictional notes, single annotator, modest sample), it provides preliminary evidence of dataset utility. Formal validation on real GP documentation with dual-annotator gold standards is planned pending ethics approval (Section~7.5).

\subsubsection{Training}
MedCAT was trained self-supervised for 0--4 epochs on SynGP500.

\subsubsection{Evaluation Metrics}
We report performance using four matching strategies of increasing clinical relevance:
\begin{itemize}[noitemsep]
    \item \textbf{Strict F1:} Exact span boundaries + exact SNOMED CUI match (most stringent, requires perfect concept identification)
    \item \textbf{Type Match F1:} Exact span boundaries + semantic type match (allows different concepts within same category, e.g., any disorder)
    \item \textbf{Grouped-Type F1:} Exact span boundaries + clinically equivalent semantic types (consolidates functionally equivalent SNOMED types, e.g., ``disorder'' $\approx$ ``morphologic abnormality'')—\emph{most clinically relevant metric}
    \item \textbf{Span F1:} Exact span boundaries only (measures entity boundary detection quality regardless of concept mapping)
\end{itemize}

\textbf{Grouped-Type matching} is emphasized as the most clinically meaningful metric. In real clinical practice, SNOMED CT semantic distinctions like ``rash (morphology)'' versus ``rash (finding)'' are functionally equivalent—both refer to the same clinical entity. Grouped-Type matching better reflects how clinical concepts are actually used in healthcare settings and provides a more realistic assessment of NLP system utility (see Appendix~B for definitions).

\subsubsection{Results}

Table~\ref{tab:ner-results} presents NER performance across training epochs using four matching strategies. Results demonstrate substantial improvements from training on SynGP500, indicating the dataset's utility for training clinical NLP models.

\begin{table}[h]
\centering
\footnotesize
\setlength{\tabcolsep}{4pt}
\caption{NER performance progression (F1 scores) across training epochs on SynGP500}
\label{tab:ner-results}
\begin{tabular}{lccccccc}
\toprule
\textbf{Strategy} & Untrained & 1e & 2e & 3e & 4e & Best & Gain \\
\midrule
Strict & 0.5882 & 0.6340 & 0.6187 & 0.6222 & 0.6258 & 1e & +7.8\% \\
Type Match & 0.5980 & 0.6625 & 0.6489 & 0.6542 & 0.6560 & 1e & +10.8\% \\
Grouped Type & 0.6059 & 0.6928 & \textbf{0.6951}$^a$ & \textbf{0.6951}$^a$ & \textbf{0.6951}$^a$ & 2--4e & \textbf{+14.7\%} \\
Span Match & 0.6196 & 0.7266 & \textbf{0.7289}$^a$ & \textbf{0.7289}$^a$ & \textbf{0.7289}$^a$ & 2--4e & \textbf{+17.6\%} \\
\bottomrule
\end{tabular}
\vspace{0.1cm}

\noindent\footnotesize $^a$Identical values reflect rounding to 4 decimal places; minor variations exist at higher precision.
\end{table}

\begin{figure}[ht]
    \centering
    \includegraphics[width=0.6\linewidth]{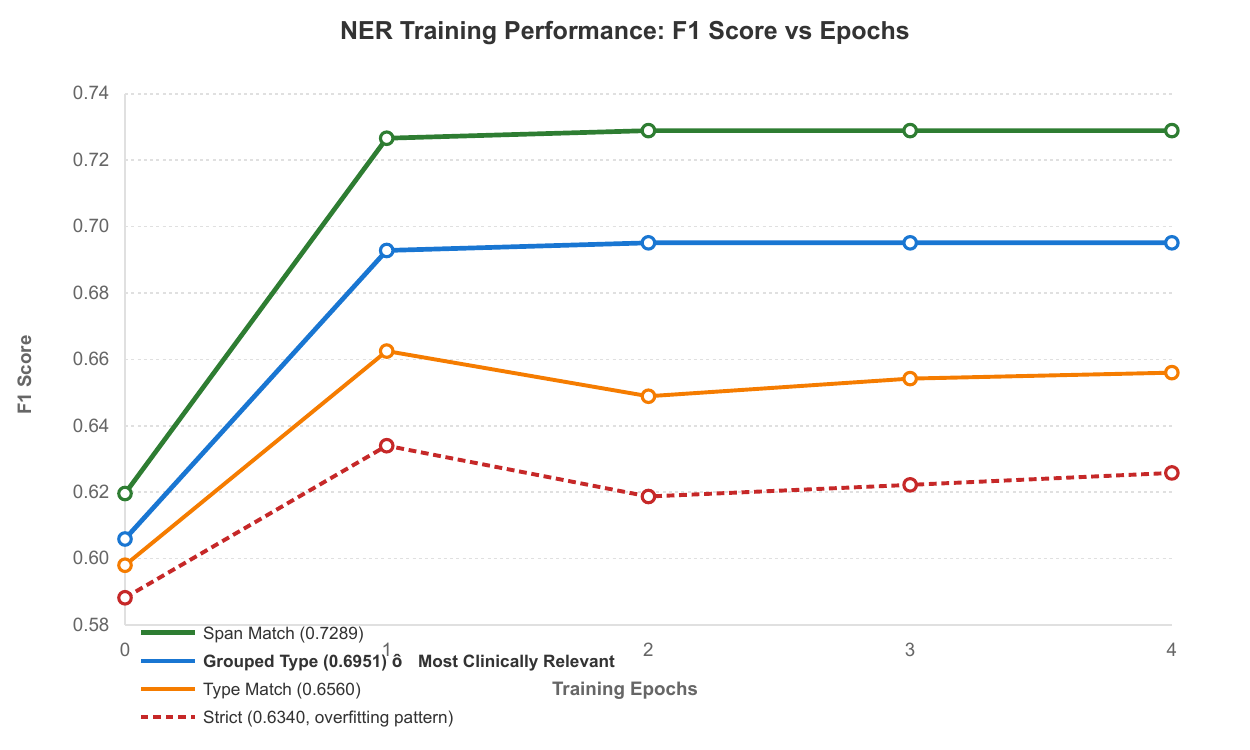}
    \caption{NER training performance across epochs on SynGP500 for multiple matching strategies. Grouped Type matching (most clinically relevant) and Span matching show strongest gains and plateau at epoch 2, while Strict matching exhibits overfitting after epoch 1.}
    \label{fig:ner-performance}
\end{figure}

Training on SynGP500 yields substantial improvements across all metrics. Grouped-Type F1 (most clinically relevant) reaches 0.6951 (+14.7\%), while Span F1 achieves 0.7289 (+17.6\%), both plateauing at 2+ epochs. These improvements demonstrate that SynGP500 provides realistic learning signal despite containing no real patient data.

\section{Illustrative Examples: Clinical Complexity in SynGP500}

Real general practice consultations rarely conform to textbook presentations. Patients juggle multiple health issues, work pressures, financial constraints, and social barriers—all competing for attention in time-limited appointments. SynGP500 captures this authentic complexity rather than sanitized, single-problem encounters. Below we present three representative examples demonstrating the dataset's realism.

\subsection{Example 1: Telegraphic Documentation Under Time Pressure}

\textbf{Case:} 24-year-old early childhood educator with impetigo (file: \path{48277006_0345_Impetigo.txt}, 29 lines)

\textbf{Clinical context:} Simple skin infection requiring occupational health documentation and work exclusion advice—demonstrates how rushed clinicians compress essential information while managing public health requirements.

\textbf{Key excerpt (lines 5--18):}
\begin{quote}
\small
\texttt{C/O 'cold sores on leg + arm' x \textasciitilde 1-2/52, itchy not painful, thinks spreading, works in childcare lots of 'school sores' around}

\texttt{Hx: Onset vague, started R shin small red spot → blister → yellow/honey crust. Now 3x R shin + 1x L forearm. No pus per pt, just crust. No fevers/chills, feels a bit tired only. [...] Using old unknown cream at home (?steroid/?abx, >1 yr).}

\texttt{O/E: afeb 36.8 HR 76 BP 112/70 RR 14 sats 99\% RA well}

\texttt{Skin: R ant shin 3 lesions 0.5-1.5 cm erythematous base + honey-coloured crust, superficial, mildly excoriated, NT, no surrounding cellulitis, no abscess. [...]]}
\end{quote}

\textbf{Realistic features:} Extreme abbreviation ("ECE" for early childhood educator, "C/O" for chief complaint); patient using incorrect terminology ("cold sores" for "school sores", demonstrating health literacy gaps); patient self-medicating with unidentified old medication; arrow notation ("→"); dense medical shorthand compressing consultation into 29 lines while still documenting work exclusion requirements ("exclude from childcare until $\geq$24 hrs after starting mupirocin").

\subsection{Example 2: Multiple Competing Demands in One Appointment}

\textbf{Case:} 57-year-old casual tiler with knee bursitis, uncontrolled diabetes, and Centrelink forms (file: \path{17059001_0012_Prepatellar_bursitis.txt}, 60 lines)

\textbf{Clinical context:} Three problems competing for attention—occupational knee injury, chronic disease non-adherence, and government paperwork—requiring pragmatic compromises shaped by socioeconomic reality.

\textbf{Key excerpt (lines 5--16, 49--52):}
\begin{quote}
\small
\texttt{CC: R knee swelling 'bloody knee keeps blowing up' + Centrelink/housing forms. DM overdue review.}

\texttt{[...] On knees all day, knee pads intermittent ('forget them in the ute').}

\texttt{DM2 {-}{-} last review \textasciitilde 9/12 ago. Multiple DNAs. On EMR: metformin XR 1g bd, gliclazide MR 60mg mane. Admits poor adherence, ran out 'a few weeks ago'. No home BGLs. Denies osmotic sx. Diet poor. Missed previous bloods.}

\texttt{SH (brief): casual tiler, variable income. Applying public housing.}

\texttt{[...]}

\texttt{Forms/admin}

\texttt{- Centrelink/housing forms completed today: documented dx chronic R prepatellar bursitis + T2DM.}

\texttt{- Noted: pain with kneeling/squatting/prolonged standing/stairs when flared. Still able to work but limited for heavy floor tiling, may need modified duties/rest breaks.}
\end{quote}

\textbf{Realistic features:} Colloquial Australian language ('bloody knee keeps blowing up'); frank documentation of non-adherence ("Multiple DNAs", 'ran out a few weeks ago'); socioeconomic barriers explicitly stated ("casual tiler, variable income"); GP completing Centrelink forms during clinical consultation (realistic administrative burden); pragmatic adaptation ("Pathology done on-site today to minimise DNA").

\subsection{Example 3: Patient-Clinician Disagreement and Dissatisfaction}

\textbf{Case:} 27-year-old retail worker with ankle sprain requesting MRI and opioids (file: \path{44465007_0024_Ankle_sprain.txt}, 68 lines)

\textbf{Clinical context:} Patient wants imaging and "strong painkillers" but evidence-based guidelines don't support either request—demonstrates realistic conflict resolution and honest documentation of patient dissatisfaction.

\textbf{Key excerpt (lines 11--13, 52--60):}
\begin{quote}
\small
\texttt{- 'Just need the strong ones again, my usual doc would just give it, no questions'}

\texttt{- 'No time for physio, can't afford days off'}

\texttt{- Asking for MRI: 'Why can't we just scan it and fix it faster?'}

\texttt{[...]}

\texttt{Imaging:}

\texttt{- X-ray/MRI not ordered {-}{-} Ottawa -ve, improving fxn, no red flags}

\texttt{- Explained MRI reserved if sx persist >6/52 / concern high-grade lig inj / intra-artic pathology}

\texttt{- Pt not happy but accepts 'better than nothing' plan}

\texttt{[...]}

\texttt{Work:}

\texttt{- Suggested ↓ standing, short seated breaks, better footwear [...]}

\texttt{- Offered med cert for light duties 1/52 {-}{-} pt declined ('boss will just cut my shifts')}
\end{quote}

\textbf{Realistic features:} Patient comparing doctors ('my usual doc would just give it'); common misconception that imaging "fixes" problems; work and financial barriers to recommended treatment ('No time for physio, can't afford days off'); honest documentation that patient left dissatisfied ('Pt not happy but accepts plan'); job insecurity preventing proper recovery ('boss will just cut my shifts').

\subsection{Summary}

These examples illustrate that SynGP500 captures the authentic complexity of general practice: time pressure forcing telegraphic documentation; multiple problems competing in single appointments; socioeconomic barriers shaping clinical decisions; patient non-adherence; disagreements between patients and clinicians; administrative burden during consultations; and pragmatic compromises. This complexity is essential for training NLP systems that will encounter real clinical text rather than sanitized textbook cases.

\section{Discussion}

\subsection{Framework Contributions}

The systematic integration of curriculum, epidemiology, and contextual variation represents a replicable framework for synthetic clinical data generation. Our approach aims to address the long-tail problem through three complementary strategies while providing diverse training conditions that may reduce domain shift—though empirical validation remains future work.

\textbf{Clinical breadth and framework scalability for the long-tail problem:} Professional medical curricula define the scope of presentations clinicians must competently manage, enabling training data to cover the full breadth of clinical scenarios models will encounter. This systematic approach to synthetic generation offers a conceptual advantage over real-world data collection: the capability to deliberately create targeted training examples unconstrained by natural disease prevalence. The current 500-note release focuses on common and less-common but important conditions defined by RACGP curriculum (Table~\ref{tab:coverage-tiers}). However, the framework's structure demonstrates scalability to truly rare long-tail conditions via specialty literature and curriculum—positioning this approach as a potential methodological pathway to a fundamental limitation of naturally occurring clinical datasets, though empirical validation of performance improvements on rare conditions remains necessary future work.

\textbf{Realistic prevalence through epidemiological calibration:} Epidemiological data prevents over-representation of rare conditions, ensuring models learn clinically appropriate concept frequency distributions. For SynGP500, BEACH study data weighted case selection to match real Australian GP consultation patterns, balancing curriculum-defined breadth with realistic prevalence.

\textbf{Deployment diversity through contextual variation:} Systematic variation across consultation settings, geographic constraints, clinician documentation styles, and psychosocial complexity prepares models for real-world heterogeneity rather than homogeneous training conditions.

This framework is replicable and adaptable to other clinical domains by substituting appropriate curriculum standards, epidemiological data sources, and jurisdiction-specific contextual variation, distinguishing this systematic approach from traditional approaches limited by natural disease prevalence.

\subsection{Validation of Synthetic Data Utility}

Our exploratory evaluation demonstrates that SynGP500 provides meaningful training signal for medical concept extraction in Australian general practice contexts, despite containing no real patient data. The substantial improvements in Grouped-Type F1 (+14.7\%) and Span F1 (+17.6\%) provide preliminary evidence that synthetic notes—when carefully grounded in clinical guidelines, epidemiology, and contextual complexity—can support machine learning model development for healthcare NLP applications. Semantic diversity analysis (Section~5.2) confirms broad embedding space coverage without template-driven clustering.

While this evaluation employed single-annotator labels and a modest test set, the consistent performance gains across multiple evaluation metrics likely indicate genuine learning signal, though validation on real clinical data remains necessary. The performance plateau after 2 epochs is consistent with expected self-supervised learning dynamics on a 500-note corpus. Importantly, clinically relevant metrics (Grouped-Type, Span) show the strongest and most stable gains, while strict CUI matching exhibits overfitting-like behavior (peak at epoch 1, then degradation). This pattern suggests the importance of evaluation metrics that align with real-world clinical utility rather than artificial precision requirements.

Our validation framework (Section 5) combines stylometric analysis, semantic diversity analysis, epidemiological calibration against BEACH consultation patterns, and NER evaluation on clinician-authored fictional test notes—enabling dataset release without requiring access to restricted clinical data. Comparison with real GP documentation is planned as future work (Section 7.5) pending ethics approval.

\subsection{Why ``Messy Reality'' Matters for Clinical NLP}

A critical contribution of SynGP500 is its deliberate inclusion of the authentic complexity and ``messiness'' that characterizes real general practice. Sanitized, textbook-style clinical notes—while easier to generate and perhaps more aesthetically pleasing—fail to prepare NLP systems for the reality they will encounter in clinical deployment.

Section 6 demonstrated the importance of clinical realism through detailed examples of authentic general practice complexity.

\subsection{Impact on Australian Health NLP Research}

SynGP500 addresses a critical national gap. Australia currently lacks publicly available clinical text corpora suitable for NLP research in general practice—the setting where most Australian healthcare is delivered. By providing privacy-preserving access without requiring ethics approval, SynGP500 enables PhD students, Australian researchers, and international researchers to explore clinical NLP methods, study Australian healthcare terminology and consultation patterns, and develop proof-of-concept systems before pursuing lengthy ethics applications for restricted datasets. The dataset supports benchmarking of models trained on other clinical corpora (e.g., MIMIC, i2b2) to quantify domain shift when deployed on Australian primary care text, filling a gap that has hindered research comparing international clinical NLP approaches with Australian healthcare contexts.

Educational and development applications include health informatics coursework requiring realistic clinical text for NLP assignments, medical documentation training demonstrating diverse consultation contexts and clinician writing styles, testing NER and information extraction architectures on Australian GP documentation, and prototyping SNOMED CT-AU-based clinical NLP systems. While SynGP500 cannot replace real-world clinical data for production system development or comprehensive validation, it provides an essential stepping stone that previously did not exist for Australian general practice NLP research.

\subsection{Limitations and Future Directions}

Several limitations warrant acknowledgment. First, pediatric cases were excluded from this release; future expansion will include pediatric presentations. Second, the generating LLM's training data composition is unknown; synthetic generation may recombine patterns from clinical text in its training corpus rather than creating entirely novel content. Third, synthetic generation may introduce subtle artifacts not present in genuine clinical documentation. Fourth, while the dataset exhibits realistic length variation, notes are likely slightly longer overall than typical time-pressured general practice documentation. Specifically, the dataset may under-represent extremely sparse documentation lacking meaningful signal (e.g., minimally informative one-line entries common under severe time pressure) while some complex presentations may be more detailed than rushed real-world consultations would produce. This design trades strict ecological validity in length distribution for practical utility: ensuring every note provides meaningful signal and educational value. Fifth, while LLM-assisted screening and GP review were conducted, large-scale multi-clinician validation has not been performed, meaning clinical inaccuracies may be present in some notes. Sixth, the evaluation set comprises fictional clinician-authored notes rather than real patient documentation, which may overestimate performance. Finally, at 500 notes, the dataset provides sufficient complexity for early-stage NLP development but remains modest compared to large-scale clinical corpora.

Future work could expand to pediatric presentations, incorporate temporal sequences for longitudinal NLP tasks, add structured annotations to reduce researcher annotation burden, and conduct formal external validation comparing synthetic versus real Australian GP documentation. The methodology could also be adapted to other clinical contexts such as emergency departments, specialist consultations, or allied health documentation.

A planned study (pending ethics approval) will compare models trained on larger synthetic corpus versus real GP data, employing dual-annotator gold standard evaluation to assess whether scaled synthetic data provides comparable learning signal.

\section{Conclusion}

SynGP500 provides the first publicly available Australian general practice clinical text dataset, addressing a critical gap in clinical NLP research infrastructure. Through clinician-curated synthetic note generation, the dataset enables early-stage NLP development, health informatics education, and research without requiring ethics approval or compromising patient privacy.

The generation framework integrates curriculum-based clinical breadth (RACGP 2022), epidemiologically calibrated prevalence (BEACH study), and authentic contextual variation—providing coverage of common presentations and less-common curriculum-specified conditions—together with deployment diversity across consultation settings, geographic locations, and documentation styles.

Multi-faceted validation demonstrates dataset quality. Stylometric analysis confirms high linguistic variation, semantic diversity analysis demonstrates absence of mode collapse, epidemiological comparison validates alignment with real Australian GP consultation patterns, and NER evaluation shows substantial performance improvements (Grouped-Type F1: +14.7\%, Span F1: +17.6\%), indicating practical utility for training machine learning models.

By deliberately capturing the authentic complexity of general practice—time pressure, patient non-adherence, socioeconomic barriers, patient-clinician disagreements, and pragmatic clinical compromises—SynGP500 prepares NLP systems for real-world deployment rather than sanitized textbook scenarios.

SynGP500 is publicly released under CC BY-NC-SA 4.0 license at \url{https://github.com/pisong314/syngp500}.

\section*{Acknowledgments}

The author acknowledges Aboriginal and Torres Strait Islander peoples as the Traditional Custodians of Australia and pays respect to Elders past, present, and emerging. This dataset includes synthetic cases that represent the diversity of patients seen in Australian general practice, including Aboriginal and Torres Strait Islander peoples. While these are entirely synthetic cases, they reflect the importance of culturally safe healthcare delivery and recognition of health disparities affecting Indigenous Australians.

\section*{Funding and Conflicts of Interest}

This work was self-funded with no external funding or conflicts of interest.

\section*{AI Assistance Disclosure}

Large language models (Claude Sonnet 4.5, Anthropic) were used to assist with manuscript drafting and editing. The author(s) take full responsibility for the content, accuracy, and integrity of this work.

\section*{Ethics Statement}

This study used entirely synthetic data with no real patient information. All 500 medical notes in SynGP500 were generated using large language models and contain no data derived from actual patients. Ethics approval was not required for the synthetic dataset creation or release. The planned validation study using real general practice data is currently awaiting ethics approval.

\section*{Data Availability}

The SynGP500 dataset is publicly available at \url{https://github.com/pisong314/syngp500} under CC BY-NC-SA 4.0 license.

\clearpage

\appendix

\section{Appendix A: BEACH Epidemiological Comparison}

\begin{table}[h]
\centering
\scriptsize
\setlength{\tabcolsep}{3pt}
\caption{SynGP500 condition distribution vs BEACH real-world GP epidemiology}
\label{tab:beach-full}
\begin{tabular}{lccc}
\toprule
\textbf{Condition Category} & \textbf{BEACH (\%)} & \textbf{SynGP500 (\%)} & \textbf{Difference} \\
\midrule
Prescription & 8.8 & 7.2 & –1.6 \\
Check-up & 8.1 & 6.4 & –1.7 \\
Test results & 6.7 & 7.0 & +0.3 \\
Cough & 4.1 & 3.6 & –0.5 \\
Immunisation / vaccination & 3.3 & 3.8 & +0.5 \\
Administrative procedure & 2.5 & 3.2 & +0.7 \\
Back complaint & 2.0 & 2.6 & +0.6 \\
Rash & 1.8 & 2.6 & +0.8 \\
Throat symptom / complaint & 1.8 & 1.4 & –0.4 \\
Blood test & 1.5 & 2.2 & +0.7 \\
Fever & 1.4 & 2.4 & +1.0 \\
Depression & 1.3 & 2.6 & +1.3 \\
Abdominal pain & 1.2 & 2.2 & +1.0 \\
Upper respiratory tract infection & 1.1 & 1.0 & –0.1 \\
Headache & 1.1 & 2.0 & +0.9 \\
Skin symptom – other & 1.1 & 3.4 & +2.3 \\
Sneezing / nasal congestion & 1.0 & 1.6 & +0.6 \\
Hypertension / high blood pressure & 1.0 & 0.6 & –0.4 \\
Anxiety & 1.0 & 1.2 & +0.2 \\
Other referrals NEC & 0.9 & 1.4 & +0.5 \\
Weakness / tiredness & 0.9 & 2.2 & +1.3 \\
Knee symptom / complaint & 0.9 & 1.2 & +0.3 \\
Observation / health-education / advice / diet & 0.9 & 1.6 & +0.7 \\
Ear pain / earache & 0.8 & 1.4 & +0.6 \\
Diarrhoea & 0.7 & 0.4 & –0.3 \\
Sleep disturbance & 0.7 & 1.2 & +0.5 \\
Swelling & 0.7 & 1.8 & +1.1 \\
Other & 40.4 & 28.1 & –12.3 \\
\bottomrule
\end{tabular}
\end{table}

\section{Appendix B: Grouped-Type Semantic Equivalence}

\begin{table}[h!]
\centering
\scriptsize
\setlength{\tabcolsep}{3pt}
\caption{Grouped-Type semantic equivalence groups with clinical examples}
\label{tab:grouped-semantic-types}
\begin{tabular}{p{3.8cm}p{9.7cm}}
\toprule
\textbf{Semantic Types} & \textbf{Real Examples} \\
\midrule
disorder / finding / morphologic abnormality &
\textbf{``fracture''}: disorder vs morphologic abnormality; \textbf{``ganglion cyst''}: disorder vs morphologic abnormality; \textbf{``bleeding''}: morphologic abnormality vs finding; \textbf{``lump''}: finding vs morphologic abnormality; \textbf{``lymphoma''}: disorder vs morphologic abnormality; \textbf{``laceration''}: disorder vs morphologic abnormality; \textbf{``deformity''}: finding vs morphologic abnormality\\
\midrule
substance / product name / medicinal product / clinical drug / product / basic dose form &
\textbf{``Clotrimazole''}: substance vs medicinal product; \textbf{``lignocaine''}: substance vs medicinal product; \textbf{``endone''}: product name vs substance\\
\midrule
procedure / regime / therapy & \\
\midrule
observable entity / attribute & \\
\midrule
occupation / role & \\
\midrule
physical object / specimen & \\
\bottomrule
\end{tabular}
\end{table}

\end{document}